\documentclass[journal]{IEEEtran}
\ifCLASSINFOpdf
\usepackage[pdftex]{graphicx}
\else
\fi
\usepackage{amsfonts}

\usepackage{amsmath}
\usepackage{array}
\begin{document}
%
\title{Predictive Modeling of Flexible EHD Pumps\\ using Kolmogorov-Arnold Networks}
%
%
%

\author{Yanhong~Peng$^{1,*}$,
        Yuxin~Wang$^{2}$,
        Fangchao~Hu$^{1}$,
        Miao~He$^{1}$,
        Zebing~Mao$^{3}$,
        Xia~Huang$^{1}$
        and~Jun~Ding$^{1}$
\thanks{$^{1}$College of Mechanical Engineering, Chongqing University of Technology, Chongqing 400054, China.}
\thanks{$^{2}$Department of Mechanical Systems Engineering, Nagoya University, Tokai National Higher Education and Research, 464-8603, Japan}
\thanks{$^{3}$Faculty of Engineering, Yamaguchi University, Yamaguchi, 755-8611,
Japan}
\thanks{$^{*}$Correspondence: yhpeng@nagoya-u.jp}
        }

\maketitle

\begin{abstract}
We present a novel approach to predicting the pressure and flow rate of flexible electrohydrodynamic pumps using the Kolmogorov-Arnold Network. Inspired by the Kolmogorov-Arnold representation theorem, KAN replaces fixed activation functions with learnable spline-based activation functions, enabling it to approximate complex nonlinear functions more effectively than traditional models like Multi-Layer Perceptron and Random Forest. We evaluated KAN on a dataset of flexible EHD pump parameters and compared its performance against RF, and MLP models. KAN achieved superior predictive accuracy, with Mean Squared Errors of 12.186 and 0.012 for pressure and flow rate predictions, respectively. The symbolic formulas extracted from KAN provided insights into the nonlinear relationships between input parameters and pump performance. These findings demonstrate that KAN offers exceptional accuracy and interpretability, making it a promising alternative for predictive modeling in electrohydrodynamic pumping.
\end{abstract}


%
\IEEEpeerreviewmaketitle

\section{Introduction}
%
%
%
%
The electrohydrodynamic (EHD) pumps are devices that harness electrostatic forces to induce the movement of a dielectric fluid, offering several advantages over traditional mechanical pumps, such as the absence of moving parts, high efficiency, and low noise \cite{peng2023review}. These characteristics make EHD pumps ideal for applications in soft robotics \cite{mao2024design} and biomedical devices  \cite{enayati2011electrohydrodynamic}. However, predicting the performance of flexible EHD pumps remains a challenge due to the intricate interactions between electrical, mechanical, and fluidic fields. In previous study \cite{mao2023soft}, two machine learning models were employed—random forest (RF) and multi-layer perceptron (MLP) to predict the performance of flexible EHD pumps. This study demonstrated that MLP, a type of feedforward artificial neural network, outperformed other models in predicting the pressure and flow rate of flexible EHD pumps. However, MLP typically rely on fixed activation functions at the nodes (or "neurons"), which can constrain the model's ability to learn complex and nonlinear relationships efficiently. Moreover, MLP can require a substantial number of neurons and layers to achieve high accuracy, leading to increased computational complexity.
Kolmogorov-Arnold Networks (KAN) \cite{liu2024kan} was proposed to promise alternatives for MLP. The KAN is a machine learning model that uses learnable spline-based functions for improved approximation of complex nonlinear relationships. Inspired by the Kolmogorov-Arnold representation theorem, KAN replace fixed activation functions with learnable activation functions on the edges ("weights"). The KAN can incorporate learnable spline functions at each weight, providing greater flexibility and accuracy in learning nonlinear relationships. 

This study introduced a novel application of KAN for predicting the performance of flexible EHD pumps and providing insights into the relations of input parameters in determining the output performance of flexible EHD pumps.

\section{Methodology}
Kolmogorov-Arnold representation \( f(x) \) is expressed as a composition of inner and outer function matrices applied to input vector \( {\bf x} \), represented as:
\[ f(x) = {\bf \Phi}_{\rm out} \circ {\bf \Phi}_{\rm in} \circ {\bf x} \]
Here, \( {\bf \Phi}_{\rm in} \) is a matrix of univariate functions, denoted as:
\[ {\bf \Phi}_{\rm in}= \begin{pmatrix} \phi_{1,1}(\cdot) & \cdots & \phi_{1,n}(\cdot) \\ \vdots & & \vdots \\ \phi_{2n+1,1}(\cdot) & \cdots & \phi_{2n+1,n}(\cdot) \end{pmatrix} \]
and \( {\bf \Phi}_{\rm out} \) is a row vector of univariate functions:
\[ {\bf \Phi}_{\rm out}=\begin{pmatrix} \Phi_1(\cdot) & \cdots & \Phi_{2n+1}(\cdot)\end{pmatrix} \]
These matrices illustrate a Kolmogorov-Arnold layer, which forms the basis of the KAN by stacking such layers. A KAN is thus constructed as:

\[ {\rm KAN}({\bf x})={\bf \Phi}_{L-1}\circ\cdots \circ{\bf \Phi}_1\circ{\bf \Phi}_0\circ {\bf x} \]

KAN builds on this theoretical foundation by replacing fixed activation functions with learnable activation functions on the weights. Each weight is represented as a learnable spline function. This architectural innovation allows KAN to better capture complex, nonlinear relationships by optimizing univariate functions directly.

In this study, we employed three distinct machine learning models—MLP, RF, and KAN—to predict the pressure and flow rate of flexible EHD pumps. Each model presents unique advantages and limitations, enabling a thorough comparative analysis. The training strategies for MLP and RF are detailed in \cite{mao2023soft}. KAN were deployed to enhance the predictive accuracy for flexible EHD pumps by replacing the fixed activation functions in MLPs with learnable univariate spline functions at each connection. The dataset has 88 training samples and 10 testing samples. The pressure prediction model featured two hidden layers and utilized cubic splines (k=3). Training was conducted using the LBFGS optimizer \cite{liu1989limited}, which is well-suited for smaller datasets. Post-training, the model underwent pruning to streamline its structure while retaining accuracy, followed by re-training for further refinement (Following pruning, the model was re-trained by continuing from the last state of weights, using the same dataset to further refine its performance.). Pruning in the context of the KAN refers to the removal of less significant weights or connections within the network. This step is designed to streamline the model by eliminating unnecessary complexity, thereby enhancing its ability to generalize without losing predictive accuracy. Symbolic formula extraction was applied to approximate the learned spline functions with mathematical expressions, enhancing model interpretability. A similar approach was adopted for the flow rate prediction model, which also incorporated two hidden layers and cubic splines (k=4), followed by the same steps of pruning, refinement, and symbolic formula extraction.

The critical geometric parameters of the flexible EHD
pumps were varied systematically to obtain comprehensive
data. These parameters include:
\begin{itemize}
\item Channel Height: Three sizes were tested: 1 mm, 0.5 mm,
and 0.15 mm.
\item Overlap of Electrode Pairs: Three overlaps were studied:
8 mm, 4 mm, and 0 mm.
\item Voltage: The voltage was set in the range of 0 to 11 kV.
\item Gap between Electrodes: Four gaps were measured: 0.3
mm, 0.6 mm, 0.9 mm, and 1.2 mm.
\item Apex Angle of Electrodes: Four angles were investigated:
$\pi$, $\pi$/2, $\pi$/3, and $\pi$/6.
\end{itemize}
Each measurement was repeated three times to ensure
accuracy, and the average values were used for further analysis.

The experiments generated a comprehensive dataset of 98 samples, each containing five input features and two output features. The input features are channel height (mm), electrode overlap (mm), voltage (V), electrode gap (mm) and  apex angle (°), while the output features comprise the maximum pressure (Pa) and maximum flow rate (ml/min) of the flexible EHD pumps. These features are represented as:
$\mathbf{X} = [X_1, X_2, X_3, X_4, X_5]$.
The output vector $Y$ is defined as:
$\mathbf{Y} = [Y_1, Y_2]$,
where: $Y_1$ is the maximum pressure (Pa), and $Y_2$ is the maximum flow rate (ml/min).
Mean Squared Error (MSE), which quantifies the average squared difference between predicted and actual values, was used to compare the performance of different machine learning models.

\begin{figure*}[!t]
\begin{center}
\includegraphics[width=14cm]{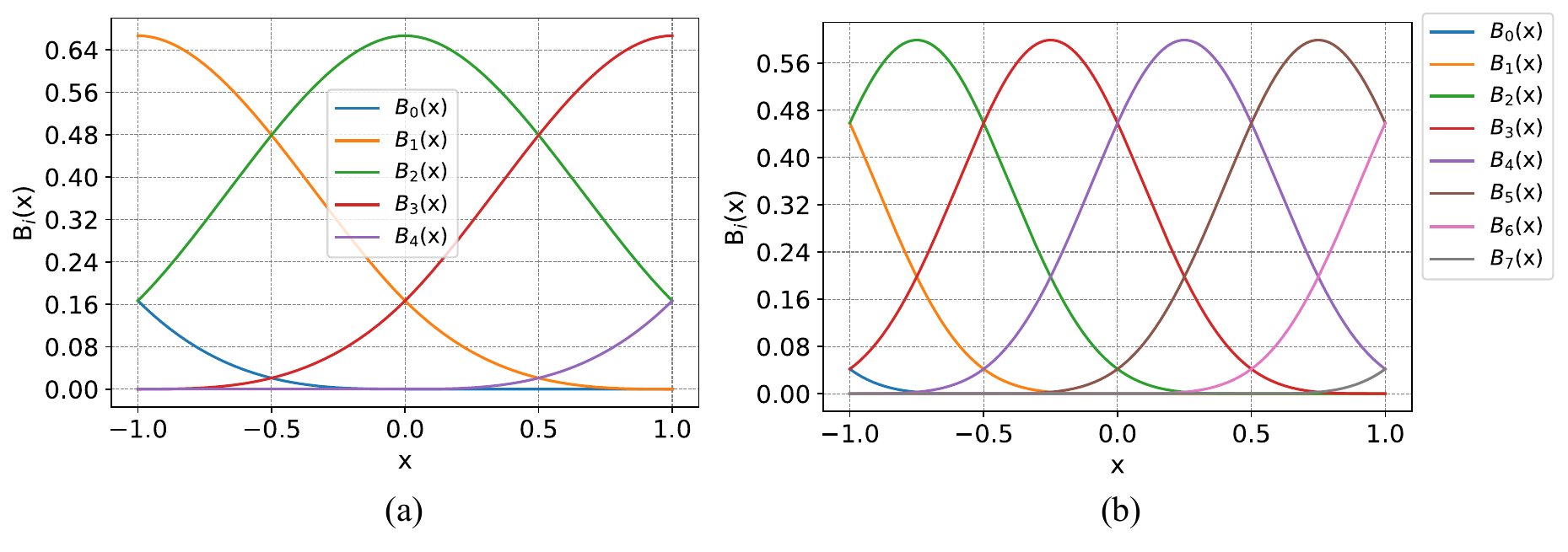}
\end{center}
\caption{Parametrize splines in (a) pressure prediction and
(b) flow rate prediction.} \label{fig:ParametrizeSplines}
\end{figure*}

\section{Result and Discussion}

After training and evaluating all models for both pressure and flow rate prediction, the results were compared comprehensively. KAN showed significant improvements in both predictive accuracy and interpretability due to its learnable activation functions and spline-based architecture.

\begin{figure*}[!t]
\begin{center}
\includegraphics[width=18cm]{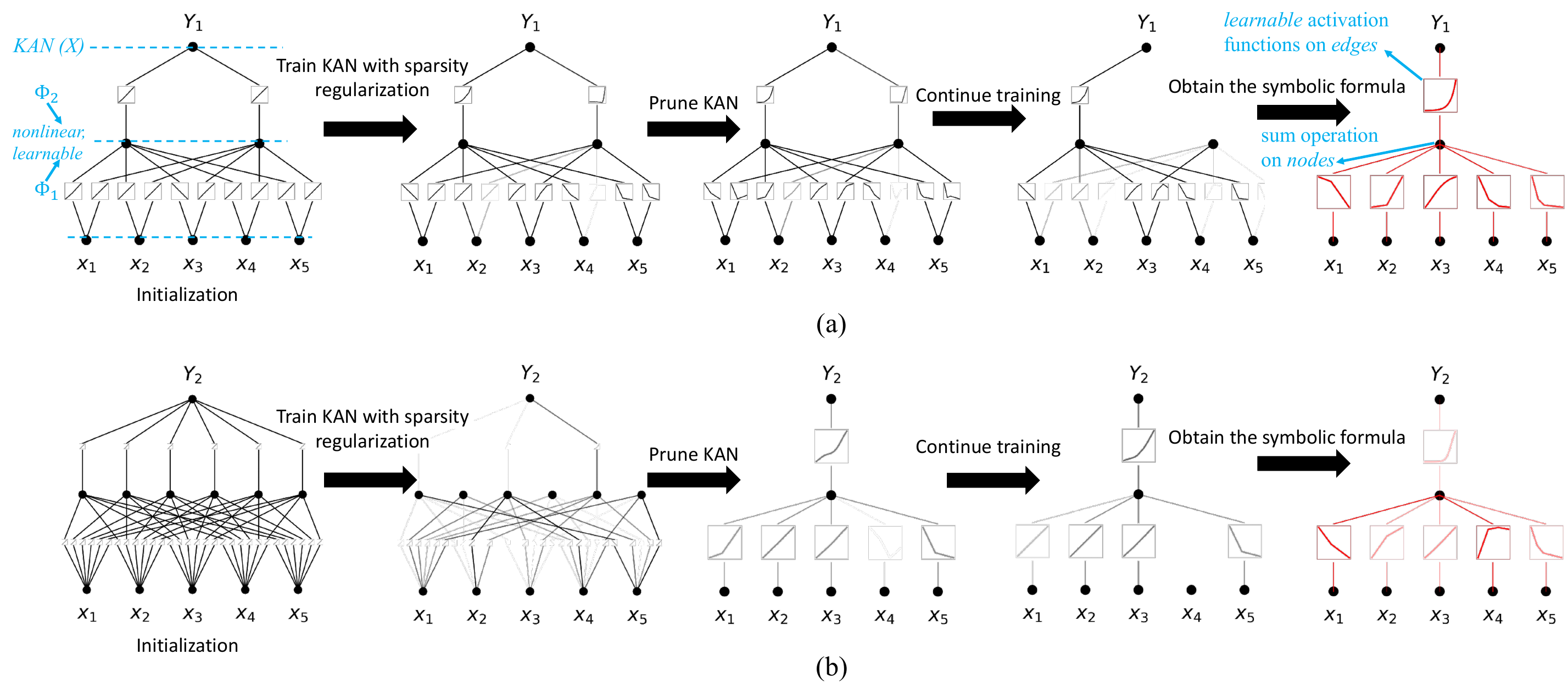}
\end{center}
\caption{Training process in (a) pressure prediction (Supplementary video S1) and (b) flow rate prediction (Supplementary video S2). The process includes: (1) Initialization of a KAN model with 5D inputs, 1D output, and hidden neurons using splines with grid intervals; (2) Initial plotting to visualize the basis functions; (3) Training with the LBFGS optimizer and sparsity regularization; (4) Pruning to simplify the model; (5) Further training and refinement; and (6) Extraction of symbolic formulas representing learned relationships.} \label{fig:Result}
\end{figure*}

Figure \ref{fig:ParametrizeSplines} (a) and (b) illustrate the basis functions ($B_i(x)$) used in the KAN model for predicting pressure and flow rate, respectively. Each function is parameterized as a B-spline, learning adaptively from input-output relationships. Figure \ref{fig:Result} (a) and (b) depict the model's training and pruning stages. Initially, the KAN's structure was established, followed by training with sparsity regularization to streamline the model. Subsequent pruning reduced the model's size, with continued training for further refinement. The process concluded with symbolic formula extraction, converting the learned spline functions into mathematical expressions, thus preserving interpretability while approximating complex nonlinear functions effectively.

KAN achieved superior predictive accuracy for pressure and flow rate prediction are depicted in Table \ref{tab:mse_comparison}. 
The symbolic formula extracted from KAN provided further interpretability, allowing us to understand the underlying relation between input features and pressure. The formula is as follows:
\begin{equation*}
\begin{split}
Y_1 &= 12.46 \exp(-0.01 (1 - 0.8 x_2)^3 + 4.3 (1 - 0.75 x_4)^4 \\& -0.06 \exp(3.03 x_1) + 3.18 \tanh(0.18 x_3 - 0.55) \\& + 2.43 \exp(-2.57 x_5)) - 1.87
\end{split}
\end{equation*}

Similarly, the formula predicting the flow rate is as follows:
\begin{equation*}
\begin{split}
Y_2 &= 1.7 - 1.59 \tanh(22.4 (0.9 - x_4)^4 \\&- 3.33 \sin(6.2 x_3 - 2.35) + 0.08 -2.11 \exp(-1.72 x_5) \\& + 2.13 \exp(-0.24 x_2) -0.89 \exp(-1.4 \cdot x_1))
\end{split}
\end{equation*}

\begin{table}[h]
\centering
\caption{MSE for pressure and flow rate predictions by different models.}
\begin{tabular}{lcc}
\hline
\textbf{Model} & \textbf{Pressure MSE} & \textbf{Flow Rate MSE} \\ \hline
KAN            & \textbf{12.186}                & 0.012                  \\ 
Random Forest  & 1750.017              & 0.040                  \\ 
MLP            & 78.329                & \textbf{0.002}                  \\ \hline
\end{tabular}
\label{tab:mse_comparison}
\end{table}

The KAN model exhibited remarkable predictive accuracy, with MSE values of 12.186 and 0.012 for pressure and flow rate predictions, respectively. These results significantly outperform Random Forest, and MLP models (in pressure prediction). The model's ability to accurately predict pressure with a low MSE makes it particularly valuable for applications in soft robotics and biomedical devices, where precise pressure control is essential. While the MLP model achieves a lower MSE for flow rate prediction, the KAN offers the distinct advantage of providing interpretable symbolic formulas that reveal the mathematical relationships between input variables and output metrics, making it particularly valuable for applications requiring a deep understanding of these underlying relationships. This formula expression has the potential to control the EHD pump by large language models in the future \cite{zhang2023large}.

\section{Conclusion}
In this study, we explored the predictive performance of the KAN in forecasting the pressure and flow rate of flexible EHD pumps. 
For pressure prediction, KAN achieved a MSE of 12.186, while in flow rate prediction, KAN obtained an MSE of 0.012. The interpretability of KAN through symbolic formula extraction provides valuable insights into the relationships between input features and output variables. The symbolic formulas reveal the significant nonlinear influence of parameters of voltage, apex angle, and electrode gap on the pressure and flow rate of EHD pumps.

\section{Declaration of competing interest}
The authors declare that they have no known competing financial interests or personal relationships that could have appeared to influence the work reported in this paper.

\section{Acknowledgments}
This work was supported by the Research Startup Fund of Chongqing University of Technology.

\ifCLASSOPTIONcaptionsoff
  \newpage
\fi

\end{document}